\documentclass[10pt,twocolumn,letterpaper]{article}

\usepackage[pagenumbers]{wacv} 

\usepackage{graphicx}
\usepackage{amsmath}
\usepackage{amssymb}
\usepackage{booktabs}

\usepackage{multirow}
\usepackage{makecell}
\usepackage{tabu}
\usepackage{xcolor}
\usepackage[font=small,skip=2pt]{caption}
\usepackage{enumitem}

\definecolor{cvprblue}{rgb}{0.21,0.49,0.74}
\usepackage[pagebackref,breaklinks,colorlinks,citecolor=cvprblue]{hyperref}

\usepackage[capitalize]{cleveref}
\crefname{section}{Sec.}{Secs.}
\Crefname{section}{Section}{Sections}
\Crefname{table}{Table}{Tables}
\crefname{table}{Tab.}{Tabs.}

\newcommand{\name}{OpenCity3D}
\newcommand{\todo}[1]{\textcolor{red}{[#1]}}

\def\wacvPaperID{*****} 
\def\confName{WACV}
\def\confYear{2025}

\begin{document}

\title{OpenCity3D: What do Vision-Language Models know about\\Urban Environments?}
\author{}
\twocolumn[{
\maketitle
\begin{center}
\vspace{-50pt}
{\large
Valentin Bieri{$^1$}, 
Marco Zamboni{$^1$}, 
Nicolas S. Blumer{$^{1, 2}$}, 
Qingxuan Chen{$^{1, 2}$}, 
Francis Engelmann{$^{1,3}$}\\}
\vspace{2mm}
$^1$ETH Zürich \quad
$^2$University of Zurich \quad
$^3$Stanford University\\
\includegraphics[width=0.9\linewidth]{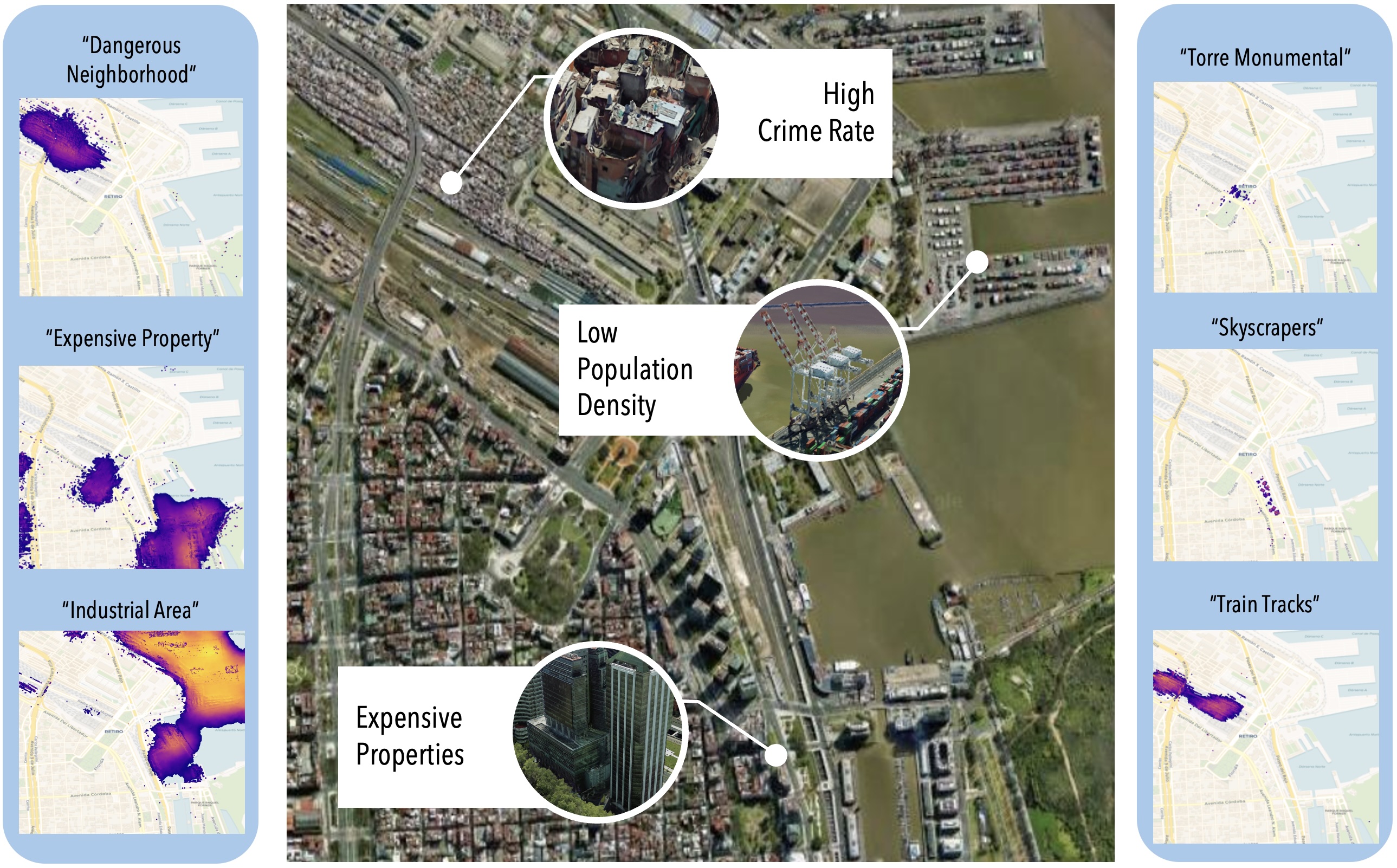}
\captionof{figure}{\name{} is a method for zero-shot urban 3D scene understanding, enabling insights into higher-level attributes such as crime rates, population density, housing prices, and local landmarks. For each text prompt, we visualize a response heatmap, where areas of higher relevance are highlighted in yellow, transitioning to blue for lower relevance.}
\label{fig:teaser}
\end{center}
}]

\begin{abstract}
\vspace{-22px}
Vision-language models (VLMs) show great promise for 3D scene understanding but are mainly applied to indoor spaces or autonomous driving, focusing on low-level tasks like segmentation.
This work expands their use to urban-scale environments by leveraging 3D reconstructions from multi-view aerial imagery.
We propose \name{}, an approach that addresses high-level tasks, such as population density estimation, building age classification, property price prediction, crime rate assessment, and noise pollution evaluation.
Our findings highlight \name{}'s impressive zero-shot and few-shot capabilities, showcasing adaptability to new contexts. This research establishes a new paradigm for language-driven urban analytics, enabling applications in planning, policy, and environmental monitoring.
See our project page: \url{opencity3d.github.io}
\vspace{-20px}\end{abstract}

\section{Introduction}
\label{sec:intro}

Recent advancements in vision-language models (VLMs) and neural rendering techniques, such as Neural Radiance Fields (NeRFs)\cite{mildenhall2020nerf} and Gaussian Splatting (GS)\cite{kerbl3Dgaussians}, have unlocked a wide range of applications in open-vocabulary 3D scene understanding.
These approaches go beyond traditional 3D scene understanding capabilities by recognizing arbitrary object classes and concepts without requiring task-specific annotated training data. Instead, they leverage the generalization power of VLMs pre-trained on internet-scale image-text pairs, which encapsulate a vast array of human knowledge and concepts.

A key application of these methods is open-vocabulary 3D segmentation, where user-provided queries -- expressed as natural language or images -- guide the recognition of regions within a 3D scene that correspond to the query.
Methods such as OpenScene~\cite{Peng2023OpenScene} and LeRF~\cite{lerf2023} have demonstrated this paradigm for 3D \emph{semantic} segmentation on polygon meshes and neural fields, respectively.
Similarly, OpenMask3D~\cite{takmaz2023openmask3d} and Open3DIS~\cite{nguyen2024open3dis} address 3D \emph{instance} segmentation.
However, these methods have been predominantly evaluated in constrained domains, such as indoor environments and autonomous driving scenarios.

This work explores, for the first time, the application of VLMs to city-scale urban 3D environments.
Understanding urban-scale properties -- ranging from building age to population density and crime rates -- is vital for urban planning and sustainable development.
Urban settings, however, introduce unique challenges, including large-scale spatial complexity and heterogeneous scene composition, which pose significant limitations for existing methods. Despite these challenges, leveraging VLMs in urban environments has the potential to provide actionable insights, offering a pathway to improved urban analysis and sustainability.

To address these challenges, we propose \name{}, a framework for open-vocabulary city-scale 3D scene understanding.
\name{} generates a language-enriched point cloud by processing RGB-D images rendered from aerial mesh reconstructions.
Through the integration of language encoders, \name{} enables querying of this enriched point cloud to analyze features of urban objects (\eg{}, buildings) and properties (\eg{}, population density, crime rates).

Our findings demonstrate significant promise for urban scene understanding, particularly in tasks such as identifying building ages, estimating housing prices, and analyzing population density, while preliminary results for higher-order tasks like crime rate prediction and noise emission analysis remain less conclusive.

In summary, our contributions are as follows:
\begin{itemize}[itemsep=0.1em, topsep=0.2em]
  \item We introduce \name{}, a novel framework for open-vocabulary city-scale 3D scene understanding.
  \item We present the first systematic analysis of VLM applicability to urban-scale question answering and scene analysis.
  \item We establish an initial benchmark for open-vocabulary urban scene understanding, providing a foundation for future research in this domain.
\end{itemize}

\begin{figure*}
\centering
\includegraphics[width=1.0\linewidth]{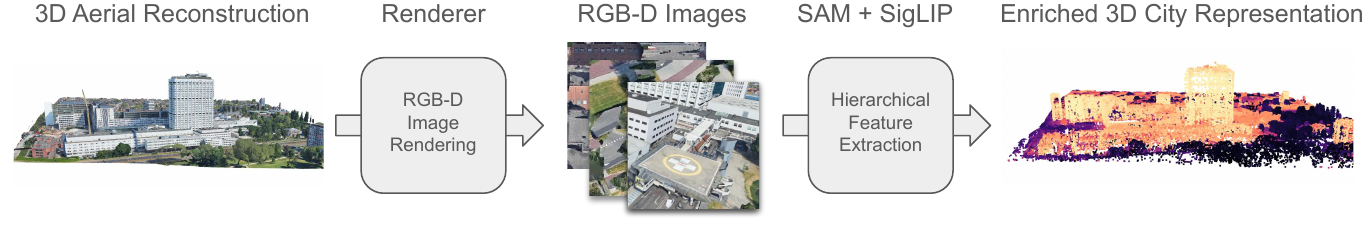}
\caption{\textbf{The \name{} model}.
Multi-view RGB-D images are rendered from aerial 3D reconstructions, followed by extracting pixel-wise hierarchical visual-language features.
These features are mapped back to the 3D mesh, enabling language-based queries.}
\label{fig:script}
\end{figure*}
\section{Related Work}

\paragraph{Open-Vocabulary 3D Segmentation}
Recent work has introduced advanced methods for 3D scene understanding, including a large variety of tasks including 3D segmentation~\cite{ji2024arkit, Peng2023OpenScene, takmaz2023openmask3d, weder2024labelmaker, yue2023agile3d}, human segmentation~\cite{human3d}, affordances~\cite{delitzas2024scenefun3d, zhang2025fungraph3d}, localization~\cite{miao2025scenegraphloc} and robot applications~\cite{lemke2024spotcompose, zurbrugg2024icgnet}. Peng~\etal{}\cite{Peng2023OpenScene} proposed OpenScene, which computes per-point features and fuses them with multi-view CLIP\cite{radford2021learning} embeddings to support open-vocabulary queries. However, OpenScene struggles with producing sharp segmentation masks and lacks instance-level discrimination.

OpenMask3D~\cite{takmaz2023openmask3d} addresses open-vocabulary 3D \emph{instance} segmentation by building on Mask3D~\cite{schult2023mask3d}, using CLIP embeddings derived from SAM~\cite{kirillov2023segment} masks on posed RGB-D images. These embeddings are matched to Mask3D instance masks, allowing comparison with text queries. A key advantage of OpenMask3D is its mask-level reasoning, which improves scalability in terms of computation and storage—crucial for city-scale 3D environments. However, as Mask3D is trained on indoor datasets, it does not generalize well to urban-scale scenes. Similarly, class-agnostic models like Segment3D~\cite{chen2022stpls3d} also underperform on city-scale data due to their indoor training regime. This concept is extended in Search3D~\cite{} to towards hierarchical 3D scenes.

LERF~\cite{lerf2023} and OpenNeRF~\cite{engelmann2024opennerf} extend NeRF~\cite{mildenhall2020nerf} to language-driven queries by learning language fields from 2D CLIP features, enabling similarity-based rendering. LangSplat~\cite{qin2023langsplat} instead combines 3D Gaussian Splatting with CLIP and SAM features, compressing VLM embeddings via an autoencoder and optimizing them through rendered-view comparisons with CLIP.

While LangSplat shows promise, scaling Gaussian Splatting to large urban environments remains challenging. Moreover, its reliance on feature compression limits open-vocabulary performance. Our approach removes this bottleneck by adapting LangSplat’s hierarchical feature extraction to a sparse point cloud representation inspired by OpenScene. By forgoing Gaussian Splatting in favor of point clouds, we retain full, uncompressed VLM features—sacrificing some geometric fidelity, which is less critical for the high-level urban analytics tasks we target.

\paragraph{Large-Scale 3D Scene Representations}
Existing methods for 3D scene understanding use diverse scene representations, including polygon meshes~\cite{Peng2023OpenScene, takmaz2023openmask3d}, point clouds~\cite{Huang2023Segment3D, takmaz2024search3d}, scene graphs~\cite{koch2024open3dsg}, NeRFs~\cite{engelmann2024opennerf}, and Gaussian splats~\cite{qin2023langsplat}.
Urban-scale environments pose unique challenges, requiring representations that scale efficiently in both memory and computation.
Implicit models such as ImpliCity~\cite{stucker2022implicity} and Block-NeRF~\cite{tancik2022blocknerfscalablelargescene} offer strong denoising and compression but incur high optimization cost and require rendering for querying.
In contrast, our method adopts an explicit point cloud representation to enable direct and efficient access to scenes.

\section{Method}

\paragraph{3D City Representation.}
\label{sec:3d_city_representation}

Fig.~\ref{fig:script} illustrates our \name{} model.
The goal is to create a 3D city representation augmented with VLM features for answering high-level socio-economical questions, such as population density, housing costs, and noise pollution.
This 3D city representation is computed in three stages:
rendering RGB-D images from a textured 3D mesh,
extracting multi-scale pixel-aligned VLM features,
and backprojecting these features onto the mesh vertices.

The input consists of a 3D polygon mesh derived from an aerial 3D reconstruction, such as Google Maps 3D Tiles~\cite{google3dtiles}.
We first render RGB color and depth images of the 3D city mesh from multiple randomly selected viewpoints.
Specifically, the camera position is randomly placed within the horizontal bounds of the 3D mesh and at a random height between $15$ m and $100$ m above sea level.
The camera azimuth orientation is sampled uniformly from
$ [0, 360^{\circ}] $,
and the elevation from $[0, 90^{\circ}]$, avoiding sky-facing views.
From the collection of rendered RGB-D images we discard images where the depth is closer than $50$ m and images where more than 20\% of the pixels are at infinite depth.

\begin{figure}
  \centering
  \begin{subfigure}{0.155\textwidth}
    \includegraphics[width=1\textwidth]{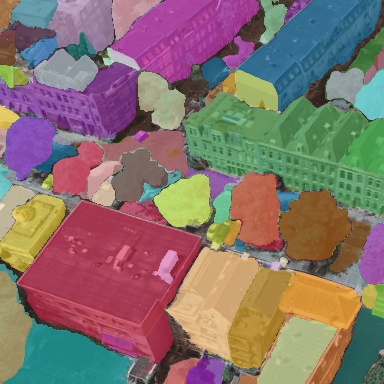}
    \caption{Large ($l$=$1$)}
    \label{fig:sam_hierarchies_large}
  \end{subfigure}
  \begin{subfigure}{0.155\textwidth}
   \includegraphics[width=1\textwidth]{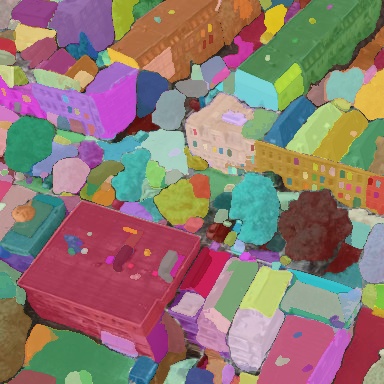}
   \caption{Medium ($l$=$2$)}
   \label{fig:short-a}
  \end{subfigure}
  \begin{subfigure}{0.155\textwidth}
    \includegraphics[width=1\textwidth]{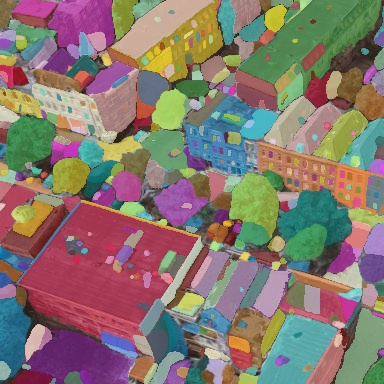}
    \caption{Small ($l$=$3$)}
    \label{fig:sam_hierarchies_small}
  \end{subfigure}
  \caption{Visualization of SAM's multi-scale segments across three hierarchy levels: small, medium, and large.}
  \label{fig:sam_hierarchies}
\end{figure}

\begin{figure}[b]
\centering
\begin{subfigure}{0.23\textwidth}
    \includegraphics[width=1\textwidth]{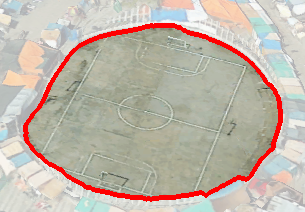}
\end{subfigure}
\begin{subfigure}{0.23\textwidth}
    \includegraphics[width=0.978\textwidth]{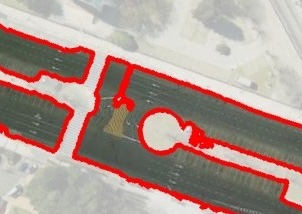}
\end{subfigure}
\caption{Example of a highlighted segments. Not removing the background provides context}
\label{fig:sam_highlighting}
\end{figure}

After rendering the RGB-D images, we apply SAM~\cite{kirillov2023segment} to the RGB images in order to obtain segments across three hierarchy levels (see Fig.~\ref{fig:sam_hierarchies}).\\
For each segment across all levels, we tightly crop the rendered image around the segment,
similar to \cite{takmaz2023openmask3d, lerf2023}, and \emph{highlight} the segmented area, as shown in Fig.~\ref{fig:sam_highlighting}.
The highlighting is a form of visual prompt tuning \cite{shtedritski2023does} which consists of two steps:
outlining the segment with a solid red line and reducing the opacity of the background.
In our experiments (Tab.~\ref{tab:compare_rotterdam}), we found that this approach outperforms existing methods such as~\cite{qin2023langsplat, liang2023open}, which completely mask the background, potentially disregarding relevant scene context.

We then process each highlighted segment with SigLIP~\cite{zhai2023sigmoid} to extract VLM features and assign them to all pixels within that segment.
We also compute the VLM feature for the entire rendered image and consider it as an additional hierarchy level: $l$=$0$.
This process results in $4$ ($3$ hierarchies from SAM and $1$ global) sets of VLM features per pixel and rendered viewpoint.

Next, we want to assign multi-level VLM features to the vertices of the mesh.
To do so we project each vertex to each of the rendered RGB-D images and, if the point is visible, assign the hierarchical VLM features of the corresponding pixel.
To verify the visibility we use the rendered depth maps to perform an occlusion test similar to \cite{Peng2023OpenScene, takmaz2023openmask3d}. 
In each hierarchy level the features of the vertex are averaged over all the images where the vertex is visible.

The point cloud defined by the vertices with the associated multi-level VLM features is the final 3D city representation that we will use for downstream tasks.
Processing a scene with $10^4$ rendered RGB images across all $L$=$4$ hierarchy levels takes 48 hours on a single NVIDIA 4090 GPU.

\paragraph{Similarity-based Prediction.}
Given a natural language text query $q$ and a set of negative queries $\mathcal{N}$, we use the SigLIP~\cite{zhai2023sigmoid} text-encoder to compute their embedding $\phi_{q}, \phi_{n}$, $n \in \mathcal{N}$.
For each point in the pointcloud we then infer the similarity scores $\overline{s}_{q, p}$ and $\overline{s}_{n, p}$ by comparing the encoded queries to the multi-level per-point features $\phi^l_{p}$ stored at point $p$ and level $l$ in the 3D scene representation.
Specifically, we follow~\cite{qin2023langsplat, lerf2023} and compute the maximal cosine similarity score $s$ across every level $l$, between the point embedding $\phi^l_{p}$ and the query embedding $\phi_{q}$:
\begin{equation}
    \overline{s}_{q, p} = \max_{l \in \{1, \cdots, L\}} \exp(\phi_{q}^\top \cdot \phi^l_{p})
  \label{eq:sim_hat}
\end{equation}
The final score for point $p$ and query $q$ is the cosine similarity described above normalized against the negative queries:
\begin{equation}
    s_{q, p} = \frac{\overline{s}_{q,p}}{\overline{s}_{q,p} + \underset{n \in \mathcal{N}}{\sum}\overline{s}_{n, p}}
  \label{eq:sim}
\end{equation}
We can use the point scores for classification (\eg{}, land use), where they serve as probabilities;
or for regression (\eg{}, property cost or building age), in which case the scores are mapped to bins:
specifically, we split the distributions of both the predicted scores and ground-truth distributions into $k$ 
quantiles (\ie bins), then map each predicted bin to the mean of the corresponding ground-truth bin.
We use $k$=$5$ for property price and building age prediction.

\paragraph{Supervised Prediction.} 
Notably, the prediction via similarity scores $s_{q,p}$ does not depend on or benefit from task-specific training data.
To overcome this we propose to perform supervised learning using the VLM features as inputs.
To do so, we discretize the ground truth values into quantile-based bins and have the classifier predict the probability of each bin.
During inference, we multiply the predicted bin probabilities with the bin centers and sum over them to obtain continuous values.
We experiment with two types of classifiers: K-Nearest Neighbors (KNN) and Light Gradient Boosting Machines (LGBMs~\cite{ke2017lightgbm}).
Unless otherwise stated, we set the number of quantiles to $k=5$ with 30\% of the data as training data and report the average over five random draws of train-test splits.

\paragraph{GPT-4o-based Prediction.}
As an alternative to SigLIP and in in order to explore the capabilities of a state of the art commercial model we also propose a method based on GPT-4o ~\cite{gpt-4o}.

For each experiment we design a specific textual prompt where we ask the model to return a value between 0 and 10 in relation to the query. In order to better guide the model we also include the string "return the result without explanation" and if possible we provide indicative numbers associated to the values (See Tab.~\ref{tab:gpt_prompts} for examples). 
We then use the RGB images rendered from the 3D mesh and feed them together with the textual prompt to the OpenAI API; in this way for each image we obtain a value between 0 and 10. In the same way as what we do with VLM features we then associate the values to the points that are visible in the image. Those values are then equivalent to $s_{q,p}$ scores and can be used for \textbf{similarity-based prediction}. 
We decided to use only full images instead of SAM segments (so only layer $l=0$) to avoid exploding costs and processing time. 

\section{Experiments}
\label{sec:experiments}

In this experiment, we assess the ability of VLMs to address broad socio-economic questions in urban settings. Using publicly available data, we evaluate six tasks across cities in the Netherlands, the US, and Buenos Aires—covering building footprint and construction year estimation (Sec.\ref{sec:netherlands}), property valuation (Sec.\ref{sec:north_am}), and socio-economic indicators such as population density, crime rates, and noise pollution (Sec.\ref{sec:buenosaires}). All experiments are conducted using VLM-augmented 3D city representations derived from texturized meshes in Google Earth\cite{google3dtiles}, as detailed in Sec.~\ref{sec:3d_city_representation}.

\subsection{The Netherlands: Building Footprint and Age}
\label{sec:netherlands}

\paragraph{Dataset.}
Our first set of experiments uses the “Basisregistraties Adressen en Gebouwen” (BAG), the official building and address registry of the Netherlands~\cite{peters2022}. This dataset provides regularly updated information on all buildings in the country; from it, we construct a benchmark of 19,349 buildings to evaluate the ability of VLMs to predict 2D building footprints and construction years.

\paragraph{Building Footprints.}
This task assesses whether VLMs can accurately predict building footprints. We formulate it as a binary classification problem, where the model labels each 3D point as either \emph{building} or \emph{background}.
As evaluation metric, we compute the \emph{Receiver Operating Characteristic Area Under the Curve}~\cite{rocauc} (ROC-AUC) score.
The ROC-AUC score indicates how clearly a classifier distinguishes positive from negative classes.
We query the 3D scene representation with the positive query \texttt{`building'} and a set of negative queries representing common urban background objects such as \texttt{`tree'} \texttt{`road'}, or \texttt{`car'} (full list in Appendix Sec.~\ref{sec:negative_prompts}).
Using Eq.~\ref{eq:sim}, we compute the similarity score $s_{q,p} \in [0,1]$ which we interpret as a probability score.
The predicted scores are projected onto a 2D grid by averaging scores of multiple 3D points per cell; empty cells are filled via linear interpolation. The resulting 2D map is then compared to ground truth footprint labels for evaluation.

We find that this classifier attains a ROC-AUC score between $86.0$\% and $94.6$\%,
accompanied by accuracies in the range of $83.2$\% and $89.8$ \%.
This is a significant improvement compared to LangSplat-style features projected to the same point cloud, which achieves only $79.8$ \% accuracy with a ROC-AUC of $86.2$ \% on the Rotterdam scene.
Furthermore, our method strongly benefits from projecting the features to a 3D point cloud, instead of directly on a 2D point grid (see Tab.~\ref{tab:2d_3d_ablation_big}, in the Appendix). 

\paragraph{Construction Year.}

In this task, the goal is to predict the year of construction of a building.
In a first experiment, we predict age scores by comparing the positive prompt \texttt{`modern building'} to the negative \texttt{`old building'}.
The ratio (Eq.~\ref{eq:sim}) between the similarity of the two is our score for the building age. 
Then we again project the points to two dimensions and re-sampling them on a regular grid. 
Each point within a building is assigned a ground truth construction year, all other points are ignored. 

The results are displayed in Tab.~\ref{tab:Netherlands_results}.
With Spearman correlations above $50$\% for both similarity-based approaches in four out of seven cities, our model provides a solid first baseline for vision-based building age prediction.
In the classifier-based setting, we train an LGBM Classifier~\cite{ke2017lightgbm},
which predicts an actual construction year instead of a similarity score.
When trained \textit{within} cities (Tab.~\ref{tab:Netherlands_results}),
this consistently results in higher correlations, combined with robust F1 scores between $0.42$ and $0.64$.
Yet, outliers such as medieval churches lead to varying Mean Absolute Error (MAE) scores - particularly in the historic Amsterdam scene.
Similar results of experiments \textit{across} scenes are displayed in the Appendix (Tab.~\ref{tab:Netherlands_results_across}). 

\begin{figure}[t]
\centering
\includegraphics[width=0.47\textwidth]{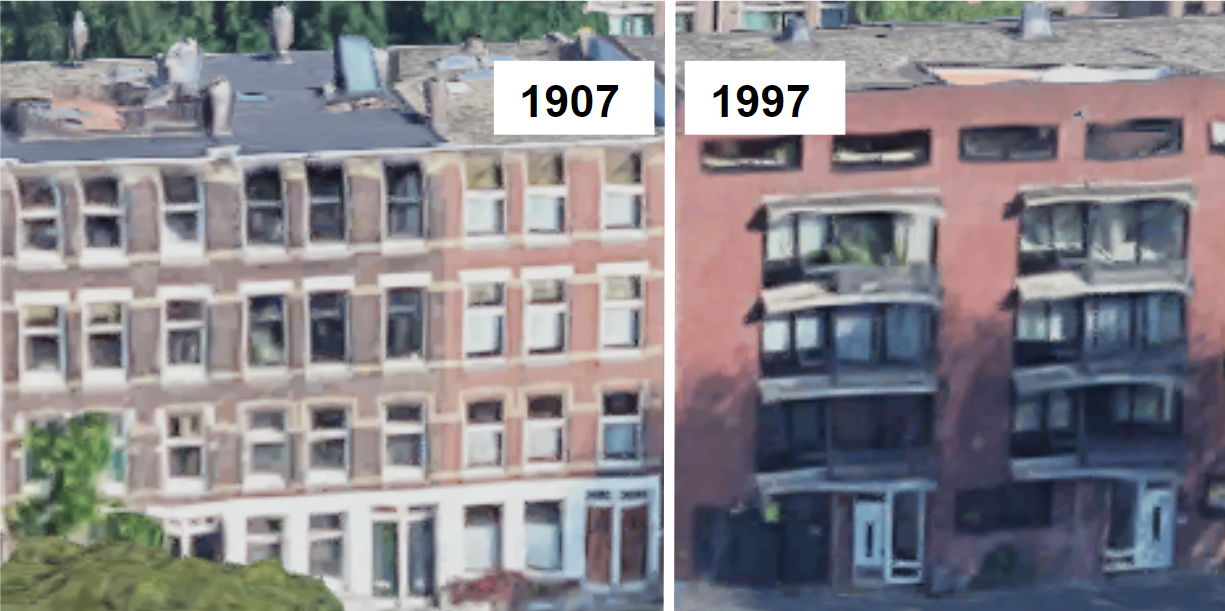}
\caption{
Illustration of the challenge to estimate building ages in the Rotterdam mesh: the left building dates from 1907, while the right one (directly adjacent) was built in 1997.
}
\label{fig:visual_age_comparison}
\end{figure}
\begin{figure*}[ht]
\vspace{20px}
\centering
\includegraphics[width=1\textwidth]{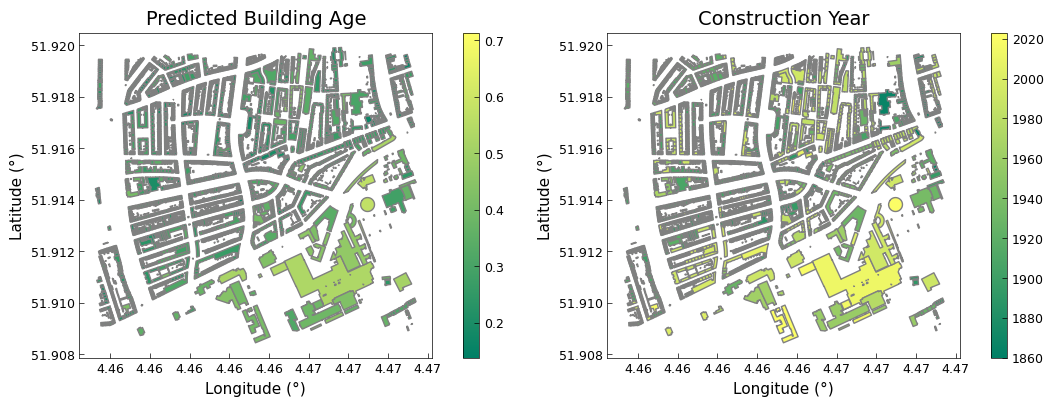}
\caption{Similarity-based predicted age \emph{(left)} vs. ground truth
construction years \emph{(right)} in Rotterdam.}
\label{fig:rotterdam_age}
\end{figure*}

Fig.~\ref{fig:rotterdam_age} shows qualitative results and illustrates how the method is able to distinguish entire districts consisting of more modern architecture from more traditional areas. Modern houses that are built back-to-back to older houses as seen in Fig.~\ref{fig:visual_age_comparison} are harder to differentiate, as these are often built to match the style of the existing neighborhood.
We furthermore find that OpenCity3D outperforms LangSplat-style features as shown in Tab.~\ref{tab:compare_rotterdam}.

\begin{table*}[ht]
    \centering
    \begin{tabular}{cccccccc}
        \toprule
        \multirow{2}{*}{City} & \multirow{2}{*}{Method} & \multicolumn{3}{c}{Building Age} & \multicolumn{3}{c}{Building Segmentation} \\
        \cmidrule(r){3-5}
        \cmidrule(r){6-8}
        & & \makecell{ Correlation} & 
        \makecell{F1 Score} &
        \makecell{MAE [y]} &  \makecell{Max Accuracy} & \makecell{ROC-AUC~\cite{rocauc}}& \makecell{F1 Score}\\
        \midrule
        \textbf{Rotterdam}         & OpenCity3D (prompt) & 0.556  & \textit{\hphantom. 0.317*} & \textit{\hphantom. 35.7*} & \textbf{87.7}\% & \textbf{0.927} & \textbf{0.796} \\
        & OpenCity3D (LGBM)& \textbf{0.769 }& \textbf{0.639} & \textbf{20.9} & 84.6\% & 0.906 & 0.702 \\
        & OpenCity3D (GPT-4o) & 0.565 & \textit{\hphantom. 0.566*} & \textit{\hphantom. 34.5*} & -- & -- & -- \\
        \midrule
        \textbf{Amsterdam} 
        & OpenCity3D (prompt) & 0.507 & \textit{\hphantom. 0.343*} & \textit{\hphantom. 240.22*} & \textbf{85.3}\% & \textbf{0.860 }& \textbf{0.722} \\
        & OpenCity3D (LGBM) & \textbf{0.577} & \textbf{0.419} & \textbf{97.7} & 76.5\% & 0.853 & 0.642 \\
        & OpenCity3D (GPT-4o) & 0.293 & \textit{\hphantom. 0.300*} & \textit{\hphantom. 251.22*} & -- & -- & -- \\
        \midrule
        \textbf{The Hague} 
        & OpenCity3D (prompt) & 0.533 & \textit{\hphantom. 0.321*} & \textit{\hphantom. 151.34*} & 83.8\% & 0.925 & \textbf{0.791}\\
        & OpenCity3D (LGBM) & \textbf{0.689} & \textbf{0.457} & \textbf{50.0} & \textbf{86.4}\% & \textbf{0.928} & 0.761\\
        & OpenCity3D (GPT-4o) & 0.498 & 0.313 & 146.80 & -- & -- & -- \\
        \midrule
        \textbf{Utrecht} 
        & OpenCity3D (prompt) & 0.364 & \textit{\hphantom. 0.280*} & \textit{\hphantom. 170.76*} & \textbf{83.2}\% & \textbf{0.866} & \textbf{0.752}\\
        & OpenCity3D (LGBM) & \textbf{0.516} & \textbf{0.482} & \textbf{58.3} & 74.1\% & 0.813 & 0.703\\
        & OpenCity3D (GPT-4o) & 0.502 & \textit{\hphantom. 0.355*} & \textit{\hphantom. 156.76*} & -- & -- & -- \\
        \midrule
        \textbf{Eindhoven} 
        & OpenCity3D (prompt) & 0.430 & \textit{\hphantom. 0.270*} & \textit{\hphantom. 27.63*} & 87.2\% & \textbf{0.931} & \textbf{0.798}\\
        & OpenCity3D (LGBM) & \textbf{0.753} & \textbf{0.501} & \textbf{12.82} & \textbf{87.7}\% & 0.899 & 0.626 \\
        & OpenCity3D (GPT-4o) & 0.636 & \textit{\hphantom. 0.380*} & \textit{\hphantom. 21.60*} & -- & -- & -- \\
        \midrule
        \textbf{Groningen} 
        & OpenCity3D (prompt) & 0.636 & \textit{\hphantom. 0.329*} & \textit{\hphantom. 81.52*} & \textbf{89.8}\% &\textbf{ 0.946} & \textbf{0.813} \\
         & OpenCity3D (LGBM) & \textbf{0.744} & \textbf{0.503} & \textbf{19.28} & 84.4\% & 0.901 & 0.665\\
        & OpenCity3D (GPT-4o) & 0.716 & \textit{\hphantom. 0.392*} & \textit{\hphantom. 74.9*} & -- & -- & -- \\
        \midrule
        \textbf{Maastricht} 
        & OpenCity3D (prompt) & 0.473 & \textit{\hphantom. 0.332*} & \textit{\hphantom. 223.81*} & \textbf{84.5}\% & \textbf{0.901} & \textbf{0.760} \\
         & OpenCity3D (LGBM) & \textbf{0.717} & \textbf{0.542 }& \textbf{57.3} & 81.8\% & 0.889 & 0.722 \\
        & OpenCity3D (GPT-4o) & 0.527 & \textit{\hphantom. 0.341*} & \textit{\hphantom. 210.50*} & -- & -- & -- \\
        
        \bottomrule
    \end{tabular}
    \caption{Result overview for building age prediction \textit{within} various cities in the Netherlands. The asterisk (*) indicates scores estimated by matching the score with the ground truth distribution based on quantiles, which is described in the supplementary material. }
    \label{tab:Netherlands_results}
\end{table*}

\begin{table}[ht]
\centering
\begin{tabular}{@{}lcc@{}}
\toprule
\makecell{Feature Type} & \makecell{Age Correlation} & \makecell{\parbox{2cm}{\centering Building seg. \\ max accuracy}}   \\
\midrule
\multicolumn{3}{l}{LangSplat} \\
\hspace{3mm}+~CLIP + prompt  & 0.394 & 79.8 \\
\hspace{3mm}+~CLIP + KNN & 0.544& 80.7\\ 
\hspace{3mm}+~SigLIP + prompt & 0.186 & 81.3\\ 
\hspace{3mm}+~SigLIP + KNN & 0.577 & 80.9 \\
\multicolumn{3}{l}{Ours} \\
\hspace{3mm}+~CLIP + prompt & 0.520 & 76.1\\
\hspace{3mm}+~CLIP + KNN & 0.681 & 79.1\\
\hspace{3mm}+~SigLIP + prompt & 0.556 & \textbf{87.7} \\
\hspace{3mm}+~SigLIP + KNN & \textbf{0.728} &  83.0 \\
\bottomrule
\end{tabular}
\caption{Evaluating feature extraction methods on the Rotterdam scene.
Mask highlighting, done by \name{}, outperforms the white-background LangSplat approach.
For LangSplat, the uncompressed, point-projected features are evaluated.}
\label{tab:compare_rotterdam}
\end{table}

\subsection{United States: Housing Prices}\label{sec:north_am}
Next, we use Zillow~\cite{zillow} data to evaluate the prediction of \emph{housing prices}.
Zillow provides a commercial analytics tool for the US real estate market that combines property data from public records with property listings from various sources. We evaluate on 1260 homes sold between 2020 and 2024 across seven US cities.

\paragraph{Housing Prices.}
Taking the point cloud with per-point features as input, we estimate the sales price of the listed homes.
To this end, we construct an indicator analogous to the previous section, using \texttt{`expensive property'} as positive and  \texttt{`cheap property'} as negative prompt.
The result is interpreted as a score for expensiveness, projected to 2D, and linearly interpolated to the known coordinates of the sold properties in the Zillow dataset.
The resulting score has correlations between $0.28$ and $0.67$ with the ground-truth sales prices. Training a LGBM Classifier \textit{across} scenes improves upon this (Tab.~\ref{tab:zillow_results}) and reaches a MAE of $0.25$M\$, which is significantly better than chance ($0.52M$\$ MAE).
These results indicate that VLMs understand some of the mechanics that determine urban property value. Their features may be a valuable addition to larger parametric models such as Zillow's \textit{Zestimate}~\cite{zestimate}.

\begin{table*}[ht]
\vspace{20px}
\centering
\resizebox{\textwidth}{!}{
\begin{tabular}{@{}lccccccccc@{}}
\toprule
& & Detroit & Miami & San Juan & Boston & San Fran. & Seattle & Los Angeles &\textbf{Overall} \\
\midrule
\multirow{3}{*}{\textbf{Spearman}} & \multicolumn{1}{l}{OpenCity3D (prompt)} & 0.528 & \textbf{0.492} & 0.348 & 0.278 & 0.674 & \textbf{0.419} & 0.504 & 0.402 \\
&\multicolumn{1}{l}{OpenCity3D (LGBM)}&   0.506 &   0.338 &   \textbf{0.432} &  \textbf{0.433} &   0.568 &   0.414 &      \textbf{0.728} &   \textbf{0.739} \\
&\multicolumn{1}{l}{OpenCity3D (GPT-4o)} & \textbf{0.600} & 0.487 & 0.260 & 0.194 & \textbf{0.710} & 0.366 & 0.192 & 0.339 \\
\midrule
\multirow{3}{*}{\textbf{F1 Score}} & \multicolumn{1}{l}{OpenCity3D (prompt)} & 0.298 & 0.278 & 0.337& 0.254 & 0.381 & 0.300 & 0.294 & 0.308 \\
&\multicolumn{1}{l}{OpenCity3D (LGBM)}   &   \textbf{0.489} &  \textbf{0.398} &   \textbf{0.340} &  \textbf{0.397} &          \textbf{0.594} &   \textbf{0.485} &      \textbf{0.541} &   \textbf{0.491} \\\
&\multicolumn{1}{l}{OpenCity3D (GPT-4o)} & 0.362 & 0.318 & 0.337 & 0.223 & 0.413 & 0.309 & 0.212 & 0.309  \\
\midrule
\multirow{3}{*}{\textbf{MAE [M\$]}} & \multicolumn{1}{l}{OpenCity3D (prompt)} & 0.201 & \textbf{0.354} & 0.477 &	0.173 & 0.179 & 0.365 & 0.276 & 0.360 \\
&\multicolumn{1}{l}{OpenCity3D (LGBM)} &
 \textbf{0.174} &  0.698 &   \textbf{0.389} &  \textbf{0.160} &  \textbf{0.163} &   \textbf{0.349} &      \textbf{0.174} &   \textbf{0.251} \\
&\multicolumn{1}{l}{OpenCity3D (GPT-4o)} & 0.195 & 0.364 & 0.498 & 0.189 & 0.171 & 0.419 & 0.350 & 0.373 \\
\bottomrule
\end{tabular}
}
\caption{Result overview for housing price prediction \textit{across scenes} in North America for zero- and few-shot setting. Zero-shot estimates of F1 and MAE are again computed by quantile-based distribution matching as described in the supplementary material. }
\label{tab:zillow_results}
\end{table*}

\subsection{Population Density, Crime Rate and Pollution}\label{sec:buenosaires}
We collect official statistics from the Autonomous City of Buenos Aires (CABA) of \textbf{population count}~\cite{bitsandbricks}, \textbf{crime records}~\cite{buenosairesdata}, and urban \textbf{noise emissions}~\cite{buenosairesdata2}. Along with them, we process one larger mesh sourced from Google Earth~\cite{google3dtiles} following Sec.~\ref{sec:3d_city_representation} to obtain a point cloud with point-wise VLM features, using the coarsest feature level.

\begin{figure*}
\vspace{20px}
  \centering
  \begin{subfigure}{0.33\linewidth}
    \includegraphics[width=7cm]{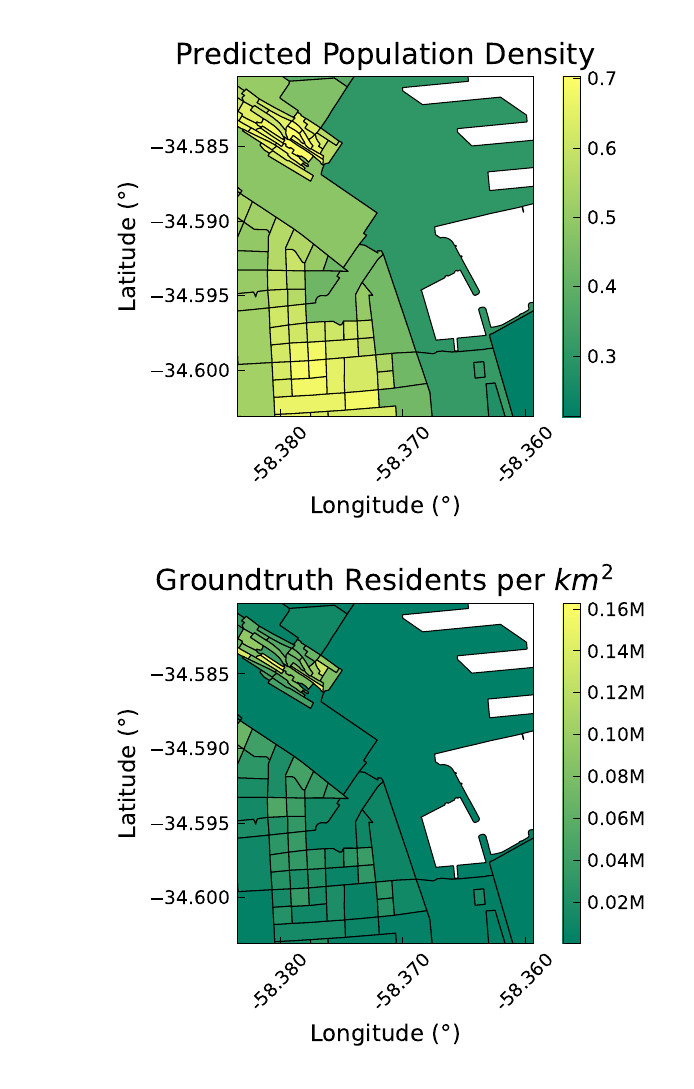}%
    \caption{Population Density.}
  \end{subfigure}
  \begin{subfigure}{0.33\linewidth}
    \includegraphics[width=7cm]{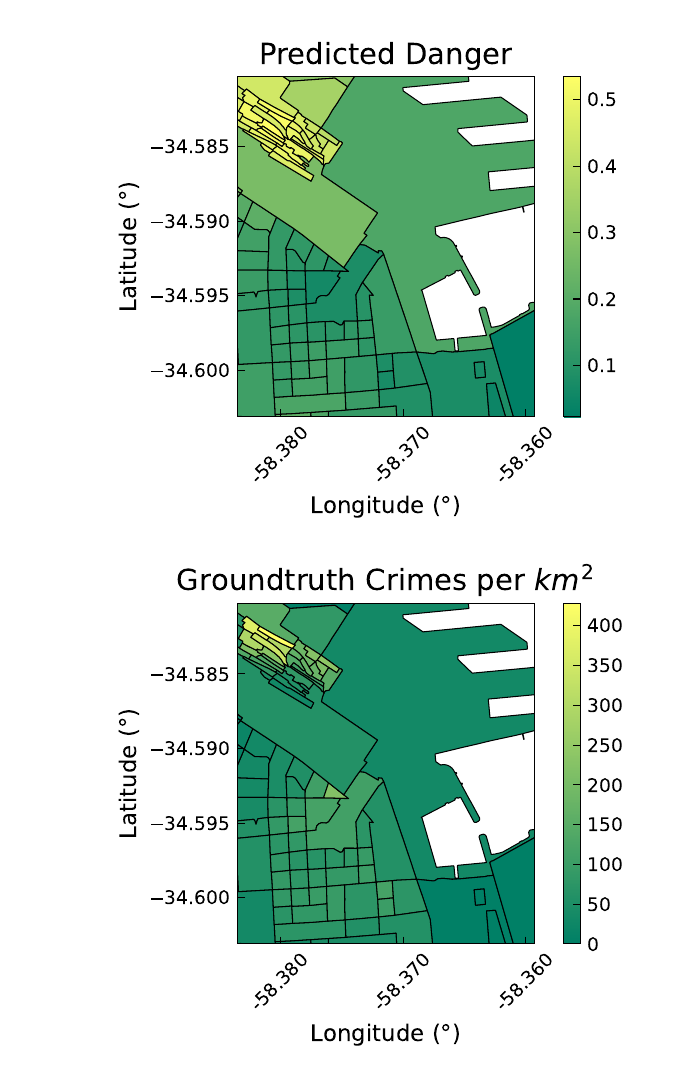}%
    \caption{Crimes per km$^2$ and year.}
  \end{subfigure}
  \begin{subfigure}{0.33\linewidth}
    \includegraphics[width=7cm]{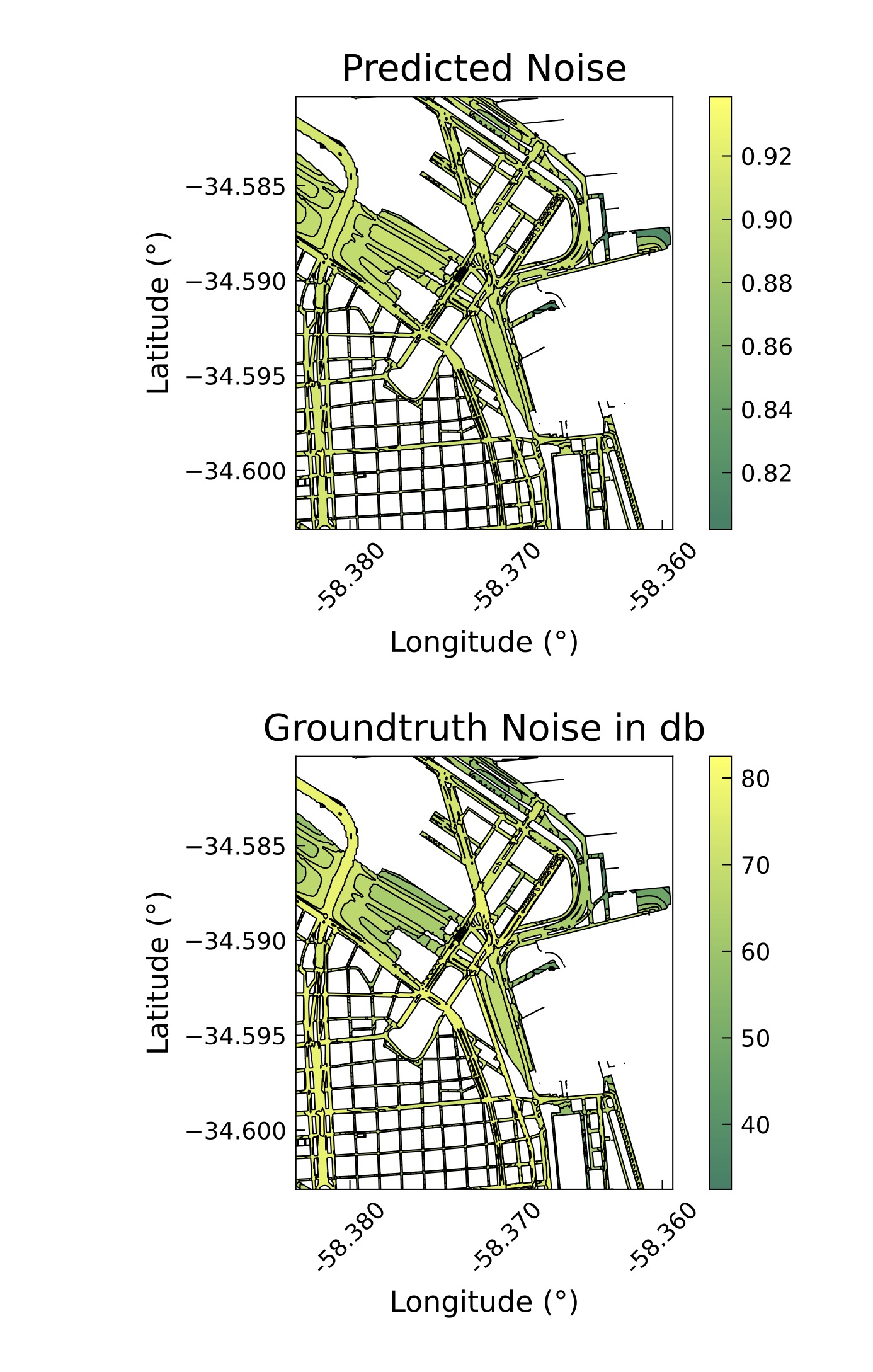}%
    \caption{Measured noise emissions.}
  \end{subfigure}
  \caption{Zero-shot prediction \emph{(top)} and ground truth \emph{(bottom)} on Buenos Aires scenes.}
  \label{fig:buenosaires_results}
\end{figure*}

\paragraph{Population Density.}
Given the point cloud and features, we use prompts to estimate population density as given by the CABA data.
The population density is given at the granularity of neighborhoods and computed by dividing the number of residents between 2015 and 2018 by the  area (see Fig.~\ref{fig:buenosaires_results}~a). We build an indicator using the positive prompts \texttt{`densely populated area'}, and \texttt{`strongly populated district'}. As negatives, we choose \texttt{`loosely populated area'}, and \texttt{`unpopulated area'}. 
Once again, we project the points to two dimensions, resample them to a regular grid, and assign them the ground truth value taken from the CABA records.
We find that the indicator yields a Spearman correlation of 0.63. 
The model correctly identifies the population cluster in the north-western section (see Fig.~\ref{fig:buenosaires_results}~a). However, it erroneously assigns high scores to the city center south of the train station. With the two additional negatives \texttt{`nature'} and \texttt{`industrial area'}, the correlation is boosted to 0.75.
We also evaluate the features in a few-shot setting, using 28 training and 94 validation neighborhoods to train a KNN regressor.
This results in a similar correlation of 0.61 (see Tab.~\ref{tab:compare_buenosaires}).
These comparably strong results do not come unexpectedly. The population density is in a direct relationship with the number and size of visible residential buildings.

\paragraph{Crime Rate.}
\label{sec:crime_rate}
Given the same features we predict Buenos Aires crime rates and validate the result against the CABA records.
CABA provides locations and descriptions of all recorded crimes between 2016 and 2022~\cite{buenosairesdata}.
We remove any crimes that do not involve a weapon to exclude incidents that are not necessarily tied to a location, such as tax evasion or fraud.
This leaves us with a dataset of 2146 crimes within the scene.
To avoid artifacts at region boundaries and attenuate sparsity effects, we consider each crime a 2D Gaussian distribution ($\sigma=50$m), from which we sample to compute the ground truth expected number of annual armed crimes per km$^2$ and neighborhood.
As an estimator we invoke Eq.~\ref{eq:sim}, using positive query \texttt{`dangerous neighborhood'} and the negative \texttt{`safe neighborhood'}. The resulting indicator obtains a relatively low Spearman correlation of $0.30$. 
As visualized in Fig.~\ref{fig:buenosaires_results}~b, the task mainly consists of identifying the port-facing side of the north-western district as a dangerous area. The model however assigns high danger scores to the port as well as the park to the southeast.
We can once again include prior knowledge to increase the correlation to $0.42$. 
In this case, however, this prior is less easily justified, as large city parks do not universally induce lower crime rates - though the mere absence of people may indicate such a tendency.
When evaluated in the aforementioned few-shot setting, KNN classification on the averaged neighborhood embeddings results in an improved correlation of 0.67. 
In summary, predicting crime rates presents itself as a complex task where many influential factors may not be immediately visible. Having reference values, like in the KNN version, greatly increased the quality
of the results. This finding indicates the need for a more nuanced approach, potentially incorporating a broader range of data types.

\paragraph{Noise Pollution.}
We follow the same procedure to estimate urban noise levels, comparing the results to official CABA measurements.
The relevant CABA noise emission dataset~\cite{buenosairesdata2} provides a map of estimated average daytime noise in decibels along major city roads (see Fig.~\ref{fig:buenosaires_results}~c).
To build an estimator, we again prompt the features using Eq.~\ref{eq:sim} with \texttt{`noisy urban area'} as positive prompt, contrasted with \texttt{`quiet area'} as a negative. This gives us a weak Spearman correlation of 0.19.
In the few-shot setting, we train a KNN regressor and obtain a moderate correlation of 0.71. Similar to the prediction of crime rates, noise level estimations remain difficult for VLMs, in particular in a zero-shot setting.

\begin{table}[ht]
\centering
\resizebox{\columnwidth}{!}{
\begin{tabular}{lccc}
\toprule
\setlength{\tabcolsep}{1pt}
Model & Population Density & Crime Rate & Noise Level \\
\midrule
Prompt & $\mathbf{62.5}$ & $42.2$ & $19.8$ \\
KNN & $60.9$ & $\mathbf{67.3}$ & $\mathbf{71.6}$\\
GPT-4o & $45.1$ & $54.4$ & $28.6$ \\
\bottomrule
\end{tabular}
}
\caption{Spearman correlations in \% for predictions: population density, crime rate, and noise levels on the Buenos Aires dataset.}
\label{tab:compare_buenosaires}
\end{table}

\section{Limitations}
A key challenge in large-scale urban 3D scene understanding is the absence of standardized datasets and benchmarks. This work takes an initial step by establishing baselines using two substantial datasets: the BAG building dataset~\cite{peters2022} and the Zillow housing dataset~\cite{zillow}. However, our dataset selection is constrained to regions where public data is available, potentially introducing a bias toward more developed areas that collect such information.

Additionally, the scale of large cities remains a technical limitation. Unlike methods such as LangSplat~\cite{qin2023langsplat}, our approach does not compress the VLM feature space into three dimensions, preserving open-vocabulary capabilities at the expense of higher memory consumption. Large cities are processed in rectangular chunks, which may introduce artifacts at chunk boundaries if overlap is insufficient. Another limitation is the reliance on relatively low-quality 3D meshes for rendering, which may affect how well VLMs interpret the imagery. To minimize bias, we recommend comparing predictions only between meshes of the same quality and source. Moreover, discrepancies between the imagery and ground-truth data may arise due to differences in capture dates (see supplementary material).

Beyond technical constraints, OpenCity3D may also reflect social and cultural biases present in visual language models. Such biases originate from the under- or over-representation of certain demographic groups in training datasets. In particular, tasks like crime rate prediction (Sec.\ref{sec:crime_rate}) risk perpetuating stereotypes and systemic discrimination, especially given the limited availability of diverse test data, as highlighted by Pouget \etal{}\cite{pouget2024no}. Approaches like those proposed by Seth \etal{}~\cite{seth2023deardebiasingvisionlanguagemodels} may help mitigate these biases.
\section{Conclusion}
This work explores whether visual-language models (VLMs) can address city-scale socio-economic questions in urban environments, including population density, building age, property value, crime rates, and noise levels.
Our experiments indicate that VLMs show strong potential for estimating population density, building age, and property value from visual data alone, while tasks like crime rate and noise level prediction remain more challenging. However, future research must account for potential biases inherent in VLMs. We view this study as a first step in this domain and hope to inspire further exploration in this direction.
{
\paragraph{Acknowledgments.}
This project originated from student coursework conducted at the CVG group at ETH Zurich and is partially supported by an ETH AI Center Postdoctoral Fellowship and an SNF Postdoc.Mobility Fellowship.
}

{\small
\bibliographystyle{ieee_fullname}
\bibliography{egbib}
}

\clearpage
\appendix
\section{Additional Results}
\subsection{Evaluation Across and Within Scenes}
\begin{figure}[b]
  \centering
  \includegraphics[width=0.85\linewidth]{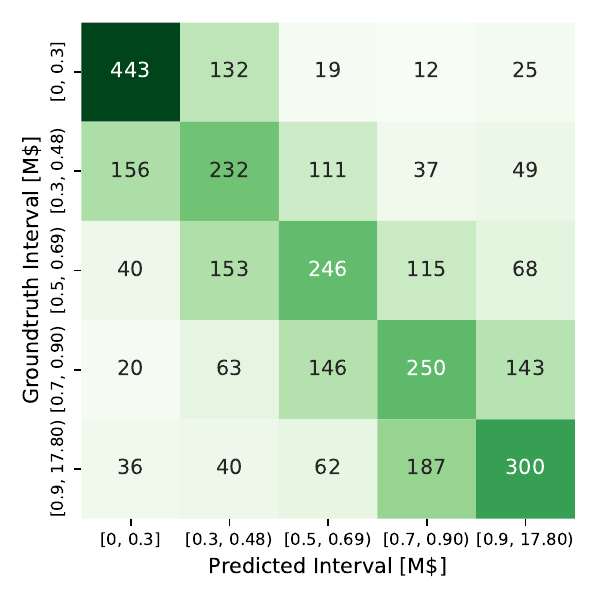}
   \caption{Confusion matrix of property price classification with LGBM~\cite{ke2017lightgbm} \textit{across} scenes.}
   \label{fig:confusion_property}
\end{figure}

\begin{figure}[b]
  \centering
  \includegraphics[width=\linewidth]{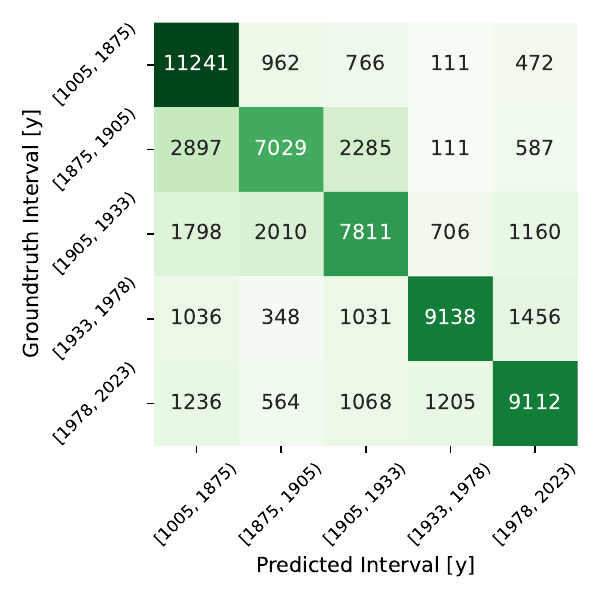}
   \caption{Confusion matrix of building age classification with LGBM~\cite{ke2017lightgbm} \textit{across} scenes.}
   \label{fig:confusion_building_age}
\end{figure}

In the main paper, we presented results for building age prediction \textit{within} cities and property price estimation \textit{across} scenes. The complementary results for building age \textit{across} cities and property price \textit{within} scenes are presented in Tables \ref{tab:Netherlands_results} and \ref{tab:property_results_within}, featuring additional metrics. Furthermore, confusion matrices are visualized in Fig.~\ref{fig:confusion_property} and Fig.~\ref{fig:confusion_building_age}.

\begin{table}[h]
\centering
\begin{tabular}{@{}lcc@{}}
\toprule
\makecell{Geometry Type} & \makecell{ROC-AUC \cite{rocauc}} & \makecell{F1 Score}   \\
\midrule
\multicolumn{3}{l}{3D Point Cloud} \\
\hspace{3mm}+ prompt & $\mathbf{0.946}$ & $\mathbf{0.813}$\\
\hspace{3mm}+ KNN & $0.828$ & $0.625$\\
\multicolumn{3}{l}{ Flat Geometry} \\
\hspace{3mm}+ prompt & $0.904$ & $0.724$ \\
\hspace{3mm}+ KNN & $0.789$ & $0.591$ \\
\bottomrule
\end{tabular}
\caption{Comparison of building segmentation performance in Groningen with a 3D point cloud vs. using a flat point grid.}
\label{tab:2d_3d_ablation_big}
\end{table}

\subsection{Evaluation with more Training Data}
\begin{figure*}[t]
  \centering
  \includegraphics[width=\linewidth]{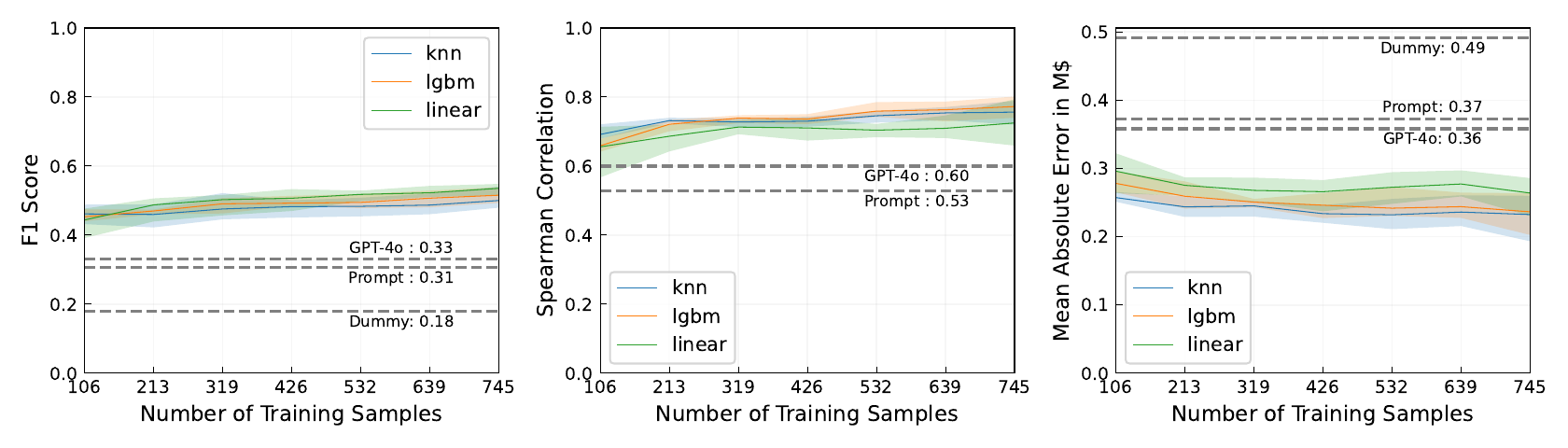}
   \caption{Property price estimation results against dataset size for experiment \textit{across} scenes. Zero-shot MAE baselines were obtained from scores by matching quantiles.}
   \label{fig:property_scale}
\end{figure*}

\begin{figure*}[t]
  \centering
  \includegraphics[width=\linewidth]{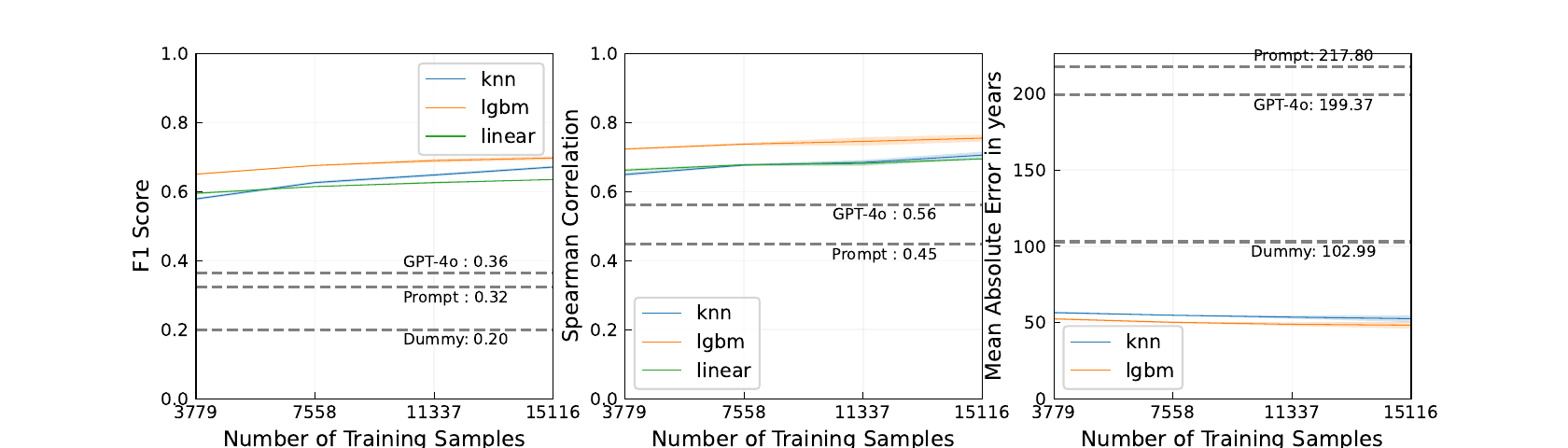}
   \caption{Building age estimation results against dataset size for experiment \textit{across} scenes. Note how quantile matching fails to produce meaningful zero-shot baselines, producing MAE significantly worse than chance.}
   \label{fig:building_age_scale}
\end{figure*}

We find that the results \textit{across} scenes can be significantly boosted when training with more than $30$\% of the dataset. Fig.~\ref{fig:property_scale} and Fig.~\ref{fig:building_age_scale} visualize this effect. 

\begin{table*}[b]
    \centering
    \begin{tabular}{lllllllll}
    \toprule
    {} & \makecell{Overall } & \makecell{Amsterdam} & \makecell{The Hague} & \makecell{Eindhoven} & \makecell{Groningen} & \makecell{Maastricht} & \makecell{Rotterdam} & \makecell{Utrecht} \\
    \multicolumn{3}{l}{F1 Score} \\
    \midrule
    lgbm   &    \textbf{0.67} &       \textbf{0.54} &       \textbf{0.47} &       \textbf{0.81} &       \textbf{0.75} &       \textbf{0.60} &       \textbf{0.76} &       \textbf{0.59} \\
    linear &    0.61 &       0.52 &       0.38 &       0.76 &       0.66 &       0.56 &       0.55 &       0.53 \\
    knn    &    0.61 &       0.51 &       0.43 &       0.78 &       0.70 &       0.54 &       0.70 &       0.49 \\
    dummy  &    0.20 &       0.23 &       0.21 &       0.28 &       0.24 &       0.21 &       0.23 &       0.21 \\
    \midrule
    \multicolumn{3}{l}{Spearman Correlation} \\
    \midrule
    lgbm   &    \textbf{0.73} &       \textbf{0.32} &       \textbf{0.56} &       \textbf{0.40} &       \textbf{0.84} &       \textbf{0.65} &       \textbf{0.76} &       \textbf{0.68} \\
    linear &    0.67 &       0.29 &       0.46 &       0.32 &       0.70 &       0.61 &       0.57 &       0.60 \\
    knn    &    0.67 &       0.25 &       0.46 &       0.33 &       0.77 &       0.56 &       0.67 &       0.52 \\
    dummy  &    0.00 &      -0.01 &       0.01 &      -0.02 &       0.03 &      -0.00 &      -0.01 &       0.01 \\
    \midrule
    \multicolumn{3}{l}{MAE [y]} \\
    \midrule
lgbm   &   \textbf{50.85} &     \textbf{122.23} &      \textbf{57.99} &      \textbf{12.64} &      \textbf{18.26} &      \textbf{63.50} &      \textbf{15.65} &      \textbf{60.62} \\
linear &   62.84 &     137.46 &      88.79 &      13.09 &      25.09 &      82.48 &      22.31 &      68.57 \\
knn    &   55.62 &     125.30 &      62.59 &      14.46 &      24.12 &      67.76 &      18.67 &      72.12 \\
dummy  &  102.95 &     166.55 &      93.28 &      77.03 &      88.49 &     106.14 &      75.75 &     109.80 \\
\midrule
 \multicolumn{3}{l}{MAPE [\%]} \\
 \midrule
lgbm   &    \textbf{3.03} &       \textbf{8.28} &       \textbf{3.11} &       \textbf{0.64} &       \textbf{0.94} &       \textbf{3.43} &       \textbf{0.81} &       \textbf{3.72} \\
linear &    3.63 &       8.94 &       4.71 &       0.66 &       1.29 &       4.40 &       1.15 &       4.10 \\
knn    &    3.30 &       8.53 &       3.36 &       0.73 &       1.23 &       3.67 &       0.96 &       4.31 \\
dummy  &    5.85 &      11.10 &       5.03 &       3.94 &       4.51 &       5.82 &       3.93 &       6.33 \\
\bottomrule
\end{tabular}
    \caption{OpenCity3D results for construction year prediction trained \textit{across} various cities in the Netherlands. }
    \label{tab:Netherlands_results_across}
\end{table*}

\begin{table*}[b]
    \centering
    \begin{tabular}{lllllllll}
    \toprule
    {} & \makecell{Mean} &\makecell{Detroit} & \makecell{Miami} & \makecell{San Juan} & \makecell{Boston} & \makecell{San Fran.} & \makecell{Seattle} & \makecell{Los Angeles}  \\
    \multicolumn{3}{l}{F1 Score} \\
    \midrule
\midrule
lgbm   &  \textbf{0.34}&  0.33 &  \textbf{0.25} &    \textbf{0.38} &   \textbf{0.34} &           0.34 &    \textbf{0.33} &       0.40  \\
linear &    0.28&  0.30 &  0.19 &    0.33 &   0.29 &           0.24 &    0.22 &       0.38  \\
knn    &   0.32&  \textbf{0.34} &  0.19 &    0.36 &   0.31 &           \textbf{0.35} &    0.26 &       \textbf{0.45}  \\
dummy  &  0.20 &  0.20 &  0.20 &    0.19 &   0.22 &           0.21 &    0.17 &       0.18 \\
    \midrule
    \multicolumn{3}{l}{Spearman Correlation} \\
    \midrule
lgbm   &  0.49&  0.55 &   0.24 &    \textbf{0.45} &   \textbf{0.49} &           0.57 &    0.39 &       0.75  \\
linear &   \textbf{0.51}& 0.55 &   \textbf{0.30} &    0.38 &   0.44 &           \textbf{0.68} &    0.43 &       \textbf{0.79}  \\
knn    &   \textbf{0.51}& \textbf{0.59} &   0.29 &    0.39 &   0.41 &           0.63 &    \textbf{0.46} &       0.77  \\
dummy  & 0.00 &   0.00 &  0.00 &    0.00 &   0.00 &          0.00 &   0.00 &      0.00 \\
    \midrule
    \multicolumn{3}{l}{MAE [M\$]} \\
\midrule
lgbm   &   0.34 & 0.19 &  1.03 &    0.39 &   \textbf{0.14} &           0.17 &    0.32 &       0.14 \\
linear &   0.37&  0.21 &  1.10 &    0.45 &   0.16 &           0.16 &    0.35 &       0.13  \\
knn    &   \textbf{0.32} & \textbf{0.17} &  \textbf{0.97} &    \textbf{0.37} &   \textbf{0.14} &           \textbf{0.14} &    \textbf{0.30} &       \textbf{0.11} \\
dummy  &     0.52& 0.28 &  1.29 &    0.55 &   0.20 &           0.39 &    0.51 &       0.39  \\
\midrule
 \multicolumn{3}{l}{RMSE [M\$]} \\
 \midrule
lgbm   &    0.58& 0.28 &  2.20 &    0.56 &   \textbf{0.17} &           0.24 &    0.42 &       0.19  \\
linear &   0.60& 0.31 &  2.23 &    0.64 &   0.21 &           0.22 &    0.44 &       0.17  \\
knn    & \textbf{0.55}&   \textbf{0.26} &  \textbf{2.15} &    \textbf{0.54} &   \textbf{0.17} &           \textbf{0.19} &    \textbf{0.39} &       \textbf{0.15}  \\
dummy  &    0.80&  0.38 &  2.60 &    0.74 &   0.25 &           0.48 &    0.64 &       0.50  \\
\bottomrule
\end{tabular}

    \caption{ OpenCity3D few-shot results for property price prediction trained \textit{within} various cities in the US. This experiment was conducted using 50\% of the samples as training data. The small training set size (down 30 samples) can otherwise lead to overfitting.}
    \label{tab:property_results_within}
\end{table*}

\subsection{Ablation: 3D Point Cloud vs. Flat Grid}
Although only evaluating on a 2D grid we find that the usage of a 3D point cloud is beneficial for feature fusing. We demonstrate that by performing an experiment without 3D geometry or depth information. That is, we project the same features onto a 2D point grid instead of the original 3D mesh.
Tab.~\ref{tab:2d_3d_ablation_big} shows that performance degrades significantly in that case. We believe that this is caused by the imprecise assignment of points to masks. An example for such imprecision are pixels near the horizon, which in this case are assumed to correspond to faraway points - although they may be occluded by the foreground.

\section{Implementation Details}
\subsection{Rendering RGB-D Views from 3D Mesh}
We sample positions based on a 2D grid, adding random offsets on all three axes.
The angle to the up-axis is sampled between 0 and 90 degrees to avoid sky-facing cameras.
The other angle is sampled uniformly at random.
RGB-D images with depth closer than $50$ m and images with infinite depth in more than $20$\% of the pixels are discarded. See Tab.~\ref{tab:example} for details on the scenes.

\subsection{Projection to Point Cloud}
The point cloud is first downsampled to $1$M points ($0.5$M if only the coarsest level was processed) to reduce memory consumption.
Following OpenMask3D~\cite{takmaz2023openmask3d}, point visibility is determined based on depth. 

However, we filter the masks before projection.
As most segments only cover a handful of pixels, we retain only those that cover at least $0.25$\% of the image.\todo{}.
This leads to the removal of roughly $60$\% of all segments and speeds up the overall processing time by 40\%.

\subsection{Prompting the Point Embeddings}
\label{sec:negative_prompts}
As mentioned in the main paper, we prompt the model with \textit{positive} and \textit{negative} queries. We find that the choice of negatives can have a strong impact on performance.
For building segmentation, the full set of negatives was: \texttt{`tree'}, \texttt{`road'}, \texttt{`park'}, \texttt{`river'}, \texttt{`car'}, \texttt{`sea / lake / canal'}, \texttt{`parking lot'}, \texttt{`urban scene'}, and \texttt{`city'}.

\subsection{Estimation}
We use scikit-learn~\cite{scikit-learn} to build unweighted KNN regressors and classifiers ($k=5$).
Each point and feature level provides a data point.
As for LightGBM\cite{ke2017lightgbm}, we use the official package with default settings.
We find that classifiers on building age, crime rate, noise levels, and population density benefit significantly from reducing noise by averaging the per-point embeddings of the relevant area (district) before training and inference. 

\subsection{Projection of Scores to Ground Truth Scale}
For property price and building age prediction, we experiment with methods to convert the scores into estimates matching the scale of the ground truth distribution. To that end, we compute the $q$ quantiles of the predicted and the ground truth distribution. Then we assign a prediction in the $i$-th quantile of the score distribution the mean of the values in the $i$-th quantile of the true distribution. We implement this strategy with $q=5$

\subsection{GPT-4o Integration}
We use GPT-4o to produce one score per prompt and image.
The obtained score is then fused into the point cloud analogously to the embeddings.
Due to cost and time constraints, we only process full images (coarsest level) and no individual segment masks from SAM.
\autoref{tab:gpt_prompts} shows the used prompts for the GPT experiments (GPT4o).
For estimating property price and building age experiments, the rating has been grounded by providing reference values for ratings 3, 6, and 9.
These reference values are obtained by binning the ground truth data into 10 bins.
Despite this grounding, the resulting scores only match the ground truth distribution to a limited extent.
We therefore evaluate them analogously to the similarity scores.
The induced prompting cost scales with the number and quality of images as well as the length of the response.
Our experiments with 7k to 10k images per scene cost 10-20\$ per query.
At the time of creation (September 2024), the inference time was roughly at 4-8h per scene.

\begin{table*}[ht]
\centering
\begin{tabular}{p{4cm}p{8cm}}
\toprule
\textbf{Experiment} & \textbf{Prompt} \\
\midrule
Noise Levels, 
Population Density and 
Dangerous Neighborhoods & Estimate the noise level, population density and how dangerous the neighborhood might be of the area shown in this image from 0 to 10. 

return the result without explanation \\
\midrule
Property Prices & Estimate the average property value of the area in the US from a scale from 0 to 10:

        3 meaning around 250k\$
        
        6 meaning around 600k\$
        
        9 meaning around 1.5m\$
        
        return the result without explanation \\
\midrule
Building Age & Estimate the average building age of the area on a scale from 0 to 10:

        3 meaning around 1739
        
        6 meaning around 1883
        
        9 meaning around 1987
        
        return the result without explanation \\
\bottomrule
\end{tabular}
\caption{GPT4-o experiments and their corresponding prompts.}
\label{tab:gpt_prompts}
\end{table*}

\subsection{Evaluation}
We evaluate our method in 2D against ground truth map data. To obtain 
Unless stated otherwise, the 3D point cloud is projected to 2D and then interpolated linearly on a regular grid. Correlation is computed on the points (not the districts/buildings). The validation set of the KNN estimators is uniformly randomly downsampled to $20$k points per scene to reduce inference time. Preliminary experiments showed that this has no significant effect on the results.

\begin{table*}
  \centering
  {\small{
  \begin{tabular}{cccccc}
    \toprule
    \makecell{Scene} & \makecell{Area (km\textsuperscript{2})} & \makecell{Latitude Bounds} & \makecell{Longitude Bounds} & \makecell{Sampling Year} & \makecell{Rendered Images} \\
    \midrule
     Buenos Aires (Argentina) & 5.20 & [-58.3801, -58.3593] & [-34.6041, -34.5803] & 2021 - 2023 & 14261\\
     Rotterdam (Netherlands) & 1.68 & [51.9088, 51.9194] & [4.4542, 4.4741] & 2019 - 2023 & 5704\\
     Amsterdam (Netherlands) & 1.99 & [52.3698, 52.3809] & [4.8937, 4.9174] & 2021 - 2023 & 6597\\
    The Hague (Netherlands) & 1.70 & [52.0782, 52.0887] & [4.3073, 4.3285] & 2020 - 2023 & 6520\\
    Utrecht (Netherlands) & 1.78 & [52.0818, 52.0929] & [5.0987, 5.1197] & 2017 - 2019 & 6527\\
    Eindhoven (Netherlands) & 1.35 & [5.42727, 5.44250] & [51.43233, 51.44241] & 2015 - 2023 & 8946\\
    Groningen (Netherlands) & 1.10 & [6.57495, 6.59036] & [53.21107, 53.21964] & 2024 & 7310\\
    Maastricht (Netherlands) & 2.20 & [5.68648, 5.70744] & [50.8425, 50.8525] & 2011 - 2023 & 12390\\
    San Juan (Puerto Rico) & 3.45 & [-66.0883, -66.0707] & [18.4475, 18.4642] & 2016 & 9369\\
    Detroit (USA) & 4.12 & [-83.0038, -82.9789] & [42.3467, 42.3648] & 2019 - 2023 & 9649\\
    Miami Beach (USA) & 3.18 & [-80.1444, -80.1272] & [25.7664, 25.7831] & 2018 - 2022 & 9377\\
    Seattle (USA) & 2.10 & [-122.39508, -122.36096] & [47.49694, 47.51248] & 2018 - 2023 & 12834\\
    Boston (USA) & 3.83 & [-70.99674, -70.96593] & [42.36831, 42.39076] & 2018 - 2021 & 14800\\
    San Francisco (USA) & 1.98 & [-122.16672, -122.15059] & [37.67978, 37.69241] & 2022 - 2023 & 9822\\
    Los Angeles & 2.67 & [-117.71718, -117.69846] & [33.61083, 33.62591] & 2017 - 2024 & 7610\\
    \bottomrule
  \end{tabular}
  }}
  \caption{Scene information. Sampling year indicates the time underlying footage for the reconstruction was taken according to Google Earth \cite{google3dtiles}.}
  \label{tab:example}
\end{table*}

\section{OpenMask3D for Urban Point Clouds}
A key characteristic of OpenMask3D~\cite{takmaz2023openmask3d} is that it segments the input point cloud and then stores one feature per 3D segment. This greatly boosts storage and memory efficiency, making it well-suited for city-scale input.

\begin{figure}[t]
  \centering
  \includegraphics[width=0.8\linewidth]{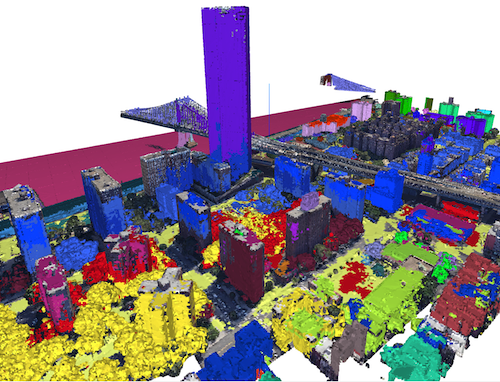}
   \caption{Example segmentation of a city area using Segment3D}
   \label{fig:segment3D}
\end{figure}

Unfortunately, OpenMask3D relies on the 3D segmentation method Mask3D~\cite{schult2023mask3d} which is trained on indoor data and therefore fails to generate meaningful segments for our 3D city scenes.
Neither OpenMask3D's Scannet200~\cite{rozenberszki2022language} and STPLS3D~\cite{chen2022stpls3d} checkpoints, nor the more recent Segment3D~\cite{Huang2023Segment3D} -- a model claimed to have superior generalization performances compared to Mask3D -- remedied the situation (see Fig.~\ref{fig:segment3D}).
In particular, we find that the models display high sensitivity to the density and scale of the point clouds. 

\begin{figure*}
        \centering
        \begin{subfigure}[b]{0.475\textwidth}
            \centering
            \includegraphics[width=0.9\textwidth]{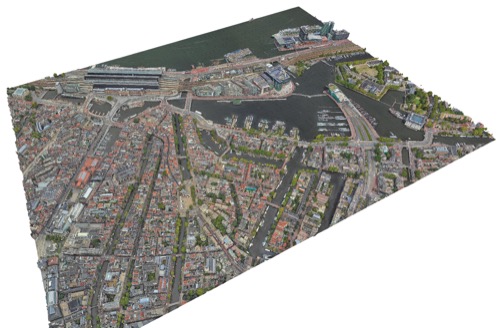}
            \caption[]%
            {{\small Rendered mesh}}    
        \end{subfigure}
        \hfill
        \begin{subfigure}[b]{0.475\textwidth}  
            \centering 
            \includegraphics[width=\textwidth]{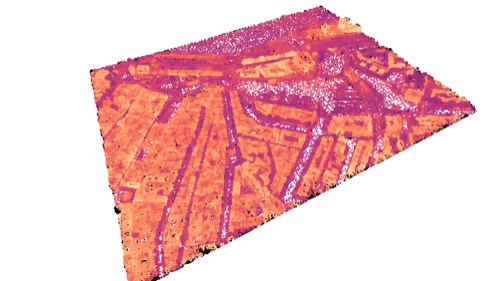}
            \caption[]%
            {{\small Prompt "building"}}    
            \label{fig:amsterdam_building}
        \end{subfigure}
        \vskip\baselineskip
        \begin{subfigure}[b]{0.475\textwidth}  
            \centering 
            \includegraphics[width=\textwidth]{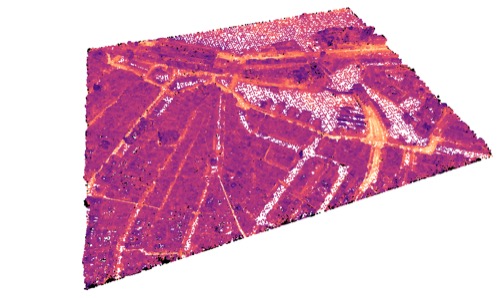}
            \caption[]%
            {{\small Prompt "road"}}    
            \label{fig:amsterdam_road}
        \end{subfigure}
        \hfill
        \begin{subfigure}[b]{0.475\textwidth}  
            \centering 
            \includegraphics[width=\textwidth]{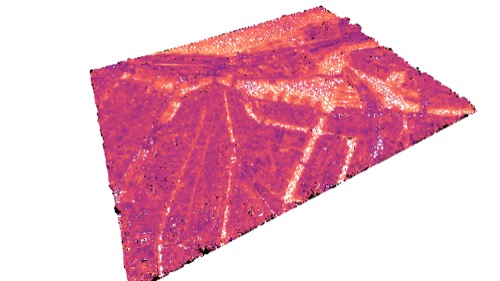}
            \caption[]%
            {{\small Prompt "water"}}    
            \label{fig:amsterdam_water}
        \end{subfigure}
        \vskip\baselineskip
        \begin{subfigure}[b]{0.475\textwidth}  
            \centering 
            \includegraphics[width=\textwidth]{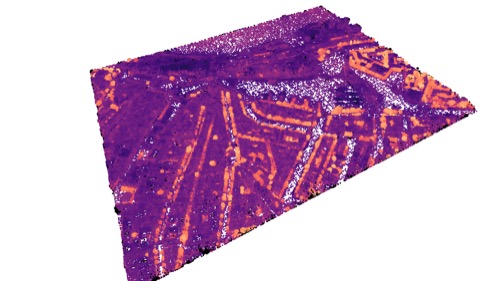}
            \caption[]%
            {{\small Prompt "tree"}}    
            \label{fig:amsterdam_tree}
        \end{subfigure}
        \hfill
        \begin{subfigure}[b]{0.475\textwidth}  
            \centering 
            \includegraphics[width=\textwidth]{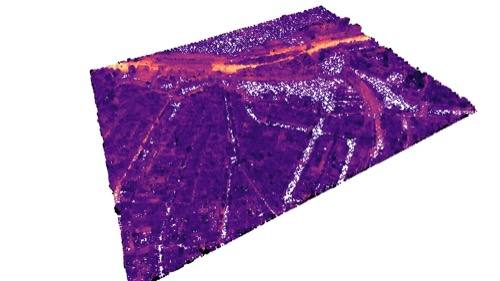}
            \caption[]%
            {{\small Prompt "train tracks"}}    
            \label{fig:amsterdam_train}
        \end{subfigure}
        \caption[]
        {\small Qualitative results for open-set segmentation in Amsterdam. We can see that buildings~\ref{fig:amsterdam_building}, trees~\ref{fig:amsterdam_tree} and train tracks~\ref{fig:amsterdam_train} are recognized with high precision, but the model has difficulties for water~\ref{fig:amsterdam_water} and roads~\ref{fig:amsterdam_road}}
        \label{fig:amsterdam_showcase}
    \end{figure*}

\section{Additional Visualizations}
\begin{figure*}[t]
  \centering
  \includegraphics[width=0.85\linewidth]{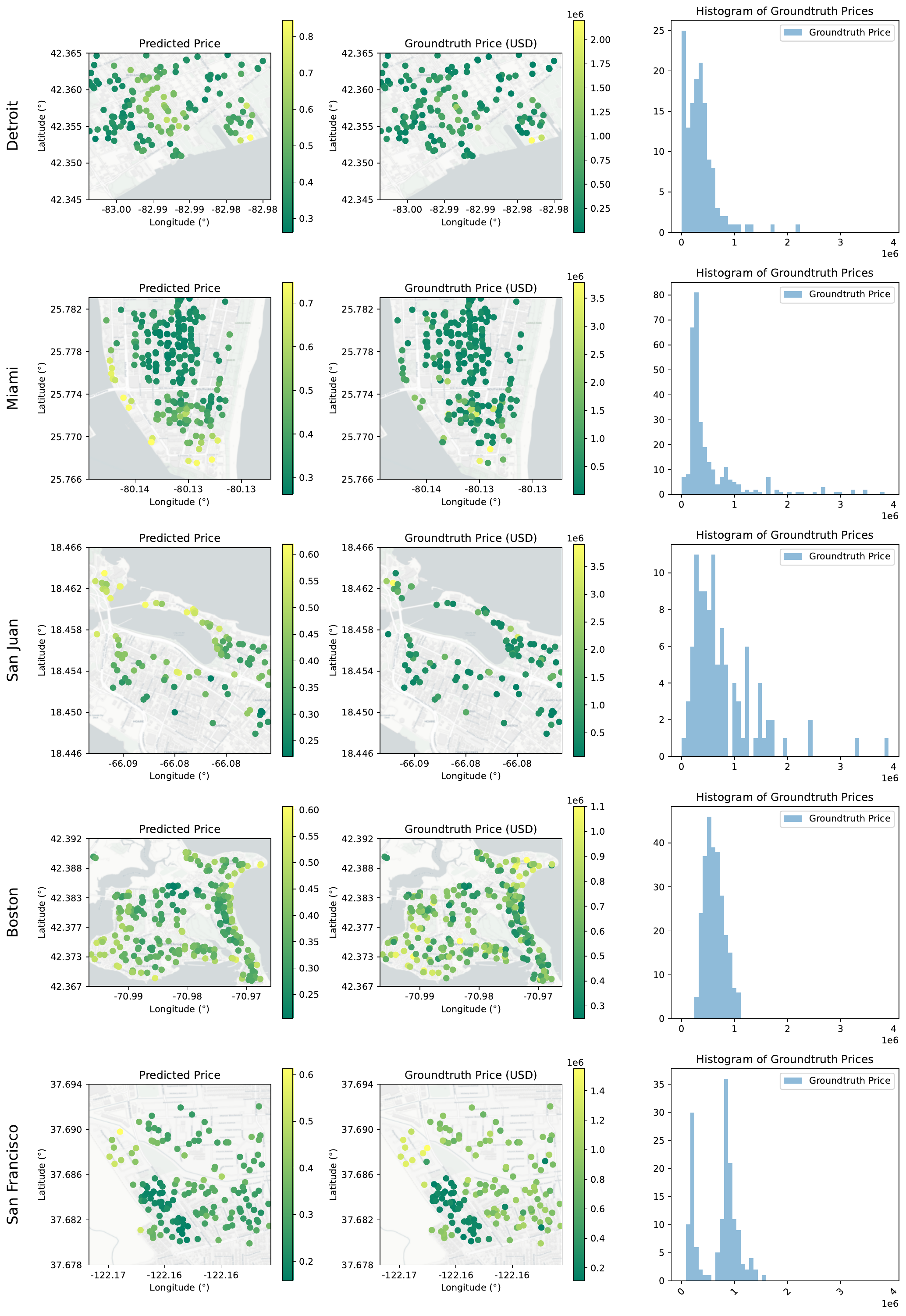}
   \caption{Visualization of zero-shot property price predictions (left) vs ground truth (right) by OpenCity. Basemaps are from CartoDB~\cite{cartodb}.}
   \label{fig:all_property_price_vis_1}
\end{figure*}

\begin{figure*}[h]
  \centering
  \includegraphics[width=0.9\linewidth]{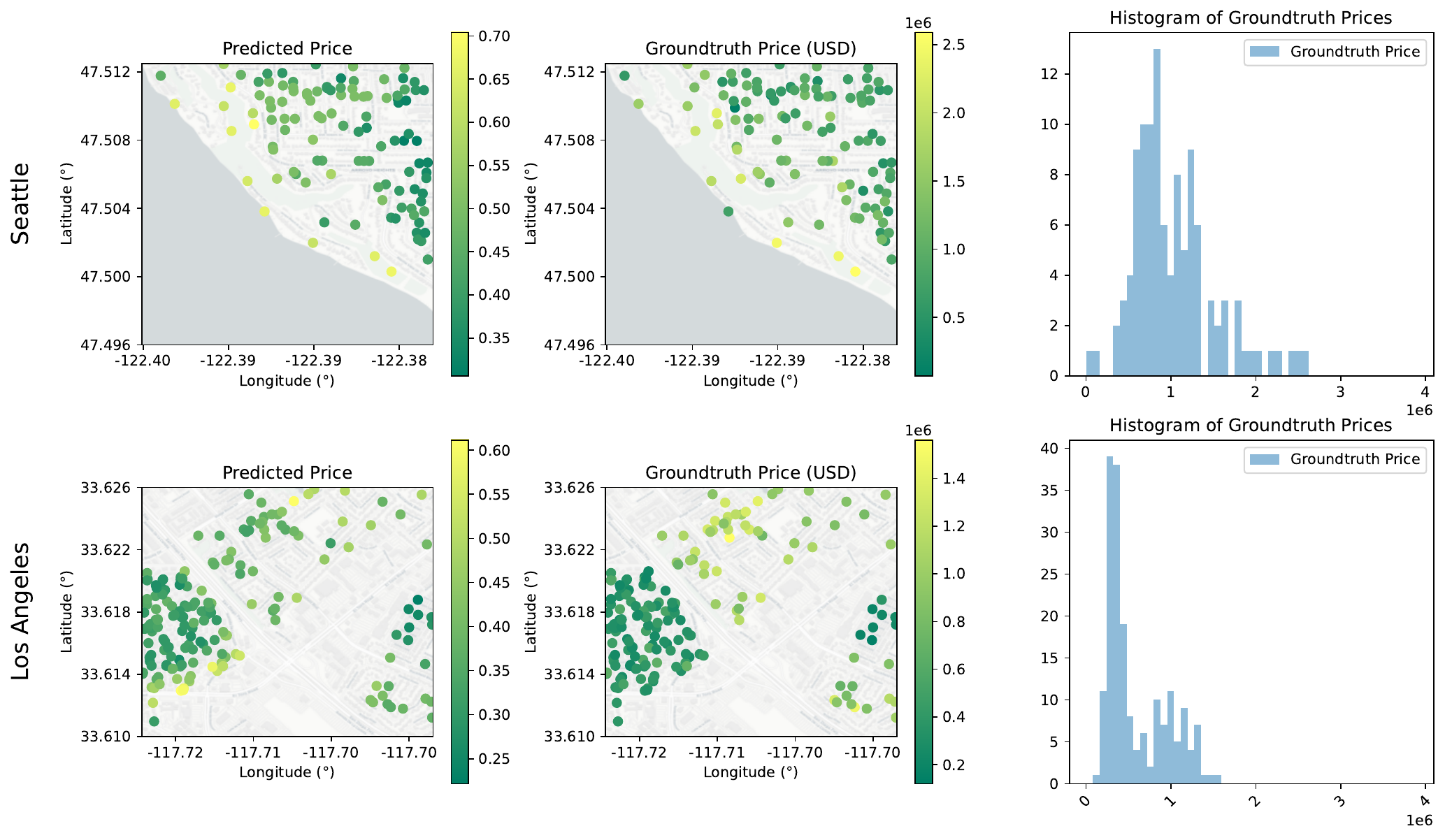}
   \caption{Visualization of zero-shot property price predictions (left) vs ground truth (right) by OpenCity. Basemaps are from CartoDB~\cite{cartodb}.}
   \label{fig:all_property_price_vis_2}
\end{figure*}

\begin{figure*}[h]
  \centering
  \includegraphics[width=0.9\linewidth]{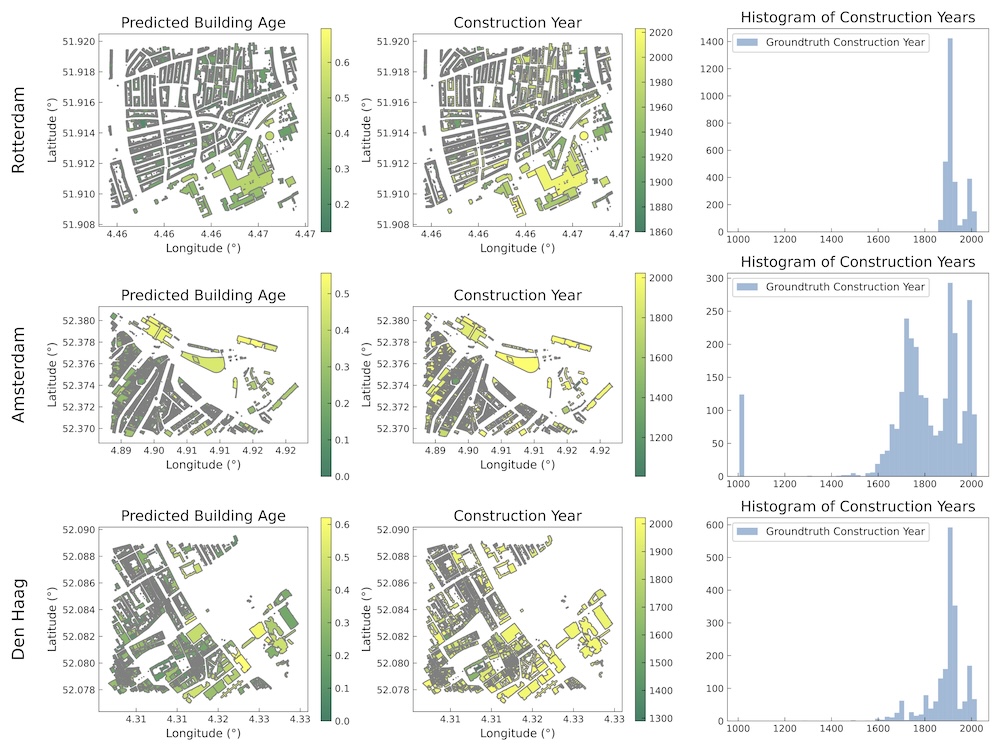}
   \caption{Visualization of zero-shot building age predictions (left) vs ground truth (right) by OpenCity. Basemaps are from CartoDB~\cite{cartodb}.}
   \label{fig:building_age_vis_1}
\end{figure*}

\begin{figure*}[h]
  \centering
  \includegraphics[width=0.9\linewidth]{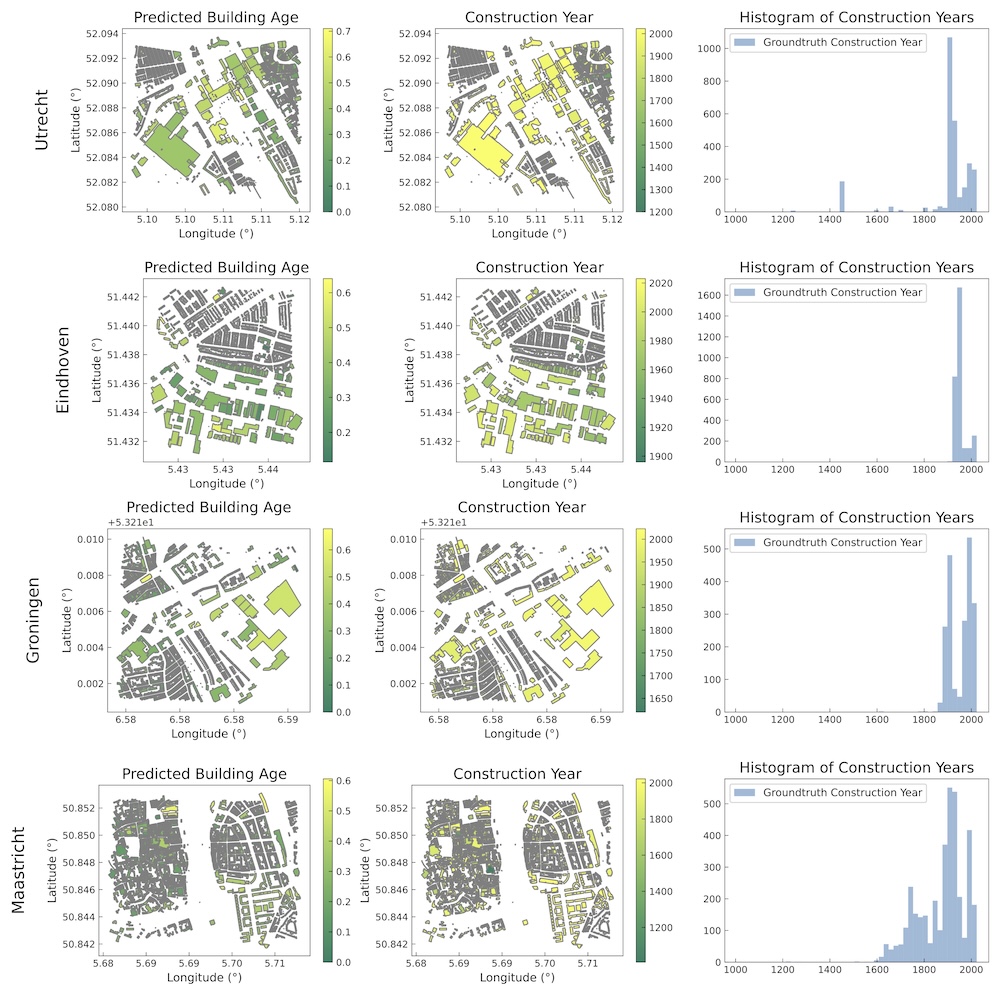}
   \caption{Visualization of zero-shot building age predictions (left) vs ground truth (right) by OpenCity. Basemaps are from CartoDB~\cite{cartodb}.}
   \label{fig:building_age_vis_2}
\end{figure*}

We provide qualitative results for open-set segmentation in Fig.~\ref{fig:amsterdam_showcase}.
Fig.~\ref{fig:all_property_price_vis_1} and Fig.~\ref{fig:all_property_price_vis_2} visualize the complete results for property price prediction, whereas Fig.~\ref{fig:building_age_vis_1} and Fig.~\ref{fig:building_age_vis_2} display the ones for building age prediction.

\end{document}